\documentclass[11pt,a4paper]{article}
\usepackage[hyperref]{acl2020}
\usepackage{times}
\usepackage{latexsym}
\usepackage{adjustbox}
\usepackage{amsfonts}
\usepackage{amsmath}
\usepackage{amssymb}

\usepackage{microtype}

\aclfinalcopy 

\newcommand\DNAME{MT-DNN}

\title{The Microsoft Toolkit of Multi-Task Deep Neural Networks for Natural Language Understanding}

\author{
Xiaodong Liu\thanks{~~Equal Contribution.}, Yu Wang$^\bold{\ast}$, Jianshu Ji, Hao Cheng, Xueyun Zhu, Emmanuel Awa,\\ \textbf{Pengcheng He, Weizhu Chen, Hoifung Poon, Guihong Cao and Jianfeng Gao}
   \\ Microsoft Corporation \\
  {\tt \{xiaodl,yuwan,jianshuj,chehao,xuzhu\}@microsoft.com}
}

\date{}

\begin{document}
\maketitle

\begin{abstract}
We present {\DNAME}\footnote{The complete name of our toolkit is $MT^2$\textit{-DNN} (The \textbf{M}icrosoft \textbf{T}oolkit of \textbf{M}ulti-\textbf{T}ask \textbf{D}eep \textbf{N}eural \textbf{N}etworks for Natural Language Understanding), but we use MT-DNN for sake of simplicity.}, an open-source natural language understanding (NLU) toolkit that makes it easy for researchers and developers to train customized deep learning models.
Built upon PyTorch and Transformers, {\DNAME} is designed to facilitate rapid customization for a broad spectrum of NLU tasks, using a variety of objectives (classification, regression, structured prediction) and text encoders (e.g., RNNs, BERT, RoBERTa, UniLM). 
A unique feature of {\DNAME} is its built-in support for robust and transferable learning using the adversarial multi-task learning paradigm.
To enable efficient production deployment, {\DNAME} supports multi-task knowledge distillation, which can substantially compress a deep neural model without significant performance drop.
We demonstrate the effectiveness of {\DNAME} on a wide range of NLU applications across general and biomedical domains. 
The software and pre-trained models will be publicly available at \hyperlink{https://github.com/namisan/mt-dnn}{https://github.com/namisan/mt-dnn}. 
\end{abstract}


\section{Introduction}
\label{sec:intro}
NLP model development has observed a paradigm shift in recent years, due to the success in using pretrained language models to improve a wide range of NLP tasks \cite{peters2018deep,devlin2018bert}. 
Unlike the traditional pipeline approach that conducts annotation in stages using primarily supervised learning, the new paradigm features a universal {\bf pretraining} stage that trains a large neural language model via self-supervision on a large unlabeled text corpus, followed by a {\bf fine-tuning} step that starts from the pretrained contextual representations and conducts supervised learning for individual tasks.
The pretrained language models can effectively model textual variations and distributional similarity. Therefore, they can make subsequent task-specific training more sample efficient and often significantly boost performance in downstream tasks. However, these models are quite large and pose significant challenges to production deployment that has stringent memory or speed requirements. 
As a result, \textbf{knowledge distillation} has become another key feature in this new learning paradigm. An effective distillation step can often substantially compress a large model for efficient deployment \cite{clark2019bam,tang2019distilling,liu2019mt-dnn-kd}.

In the NLP community, there are several well designed frameworks for research and commercial purposes, including toolkits for providing conventional layered linguistic annotations \cite{manning2014stanford}, platforms for developing novel neural models \cite{gardner2018allennlp} and systems for neural machine translation \cite{ott2019fairseq}.
However, it is hard to find an existing tool that supports all features in the new paradigm and can be easily customized for new tasks.
For example, \cite{Wolf2019HuggingFacesTS} provides a number of popular Transformer-based \cite{vaswani2017attention} text encoders in a nice unified interface, but does not offer multi-task learning or adversarial training, state-of-the-art techniques that have been shown to significantly improve performance.
Additionally, most public frameworks do not offer knowledge distillation. A notable exception is DistillBERT \cite{sanh2019distilbert}, but it provides a standalone compressed model and does not support task-specific model compression that can further improve performance.
We introduce {\DNAME}, a comprehensive and easily-configurable open-source toolkit for building robust and transferable natural language understanding models. {\DNAME} is built upon PyTorch \cite{paszke2019pytorch} and the popular Transformer-based text-encoder interface \cite{Wolf2019HuggingFacesTS}.
It supports a large inventory of pretrained models, neural architectures, and NLU tasks, and can be easily customized for new tasks. 

A key distinct feature for {\DNAME} is that it provides out-of-box adversarial training, multi-task learning, and knowledge distillation.
Users can train a set of related tasks jointly to amplify each other. They can also invoke adversarial training \cite{miyato2018virtual,jiang2019smart,liu2020alum}, which helps improve model robustness and generalizability.
For production deployment where large model size becomes a practical obstacle, users can use {\DNAME} to compress the original models into substantially smaller ones, even using a completely different architecture (e.g., compressed BERT or other Transformer-based text encoders into LSTMs \cite{hochreiter1997lstm}).
The distillation step can similarly leverage multi-task learning and adversarial training.
Users can also conduct pretraining from scratch using the masked language model objective in {\DNAME}. Moreover, in the fine-tuning step, users can incorporate this as an auxiliary task on the training text, which has been shown to improve performance.
{\DNAME} provides a comprehensive list of state-of-the-art pre-trained NLU models, together with step-by-step tutorials for using such models in general and biomedical applications.

\vspace{-0.2cm}
\section{Design}
\label{sec:design}
\begin{figure}[ht]
	\centering
	 \adjustbox{trim={0.08\width} {0.18\height} {0.02\width} {0.1\height},clip}	
     {
 	\includegraphics[width=1\textwidth]{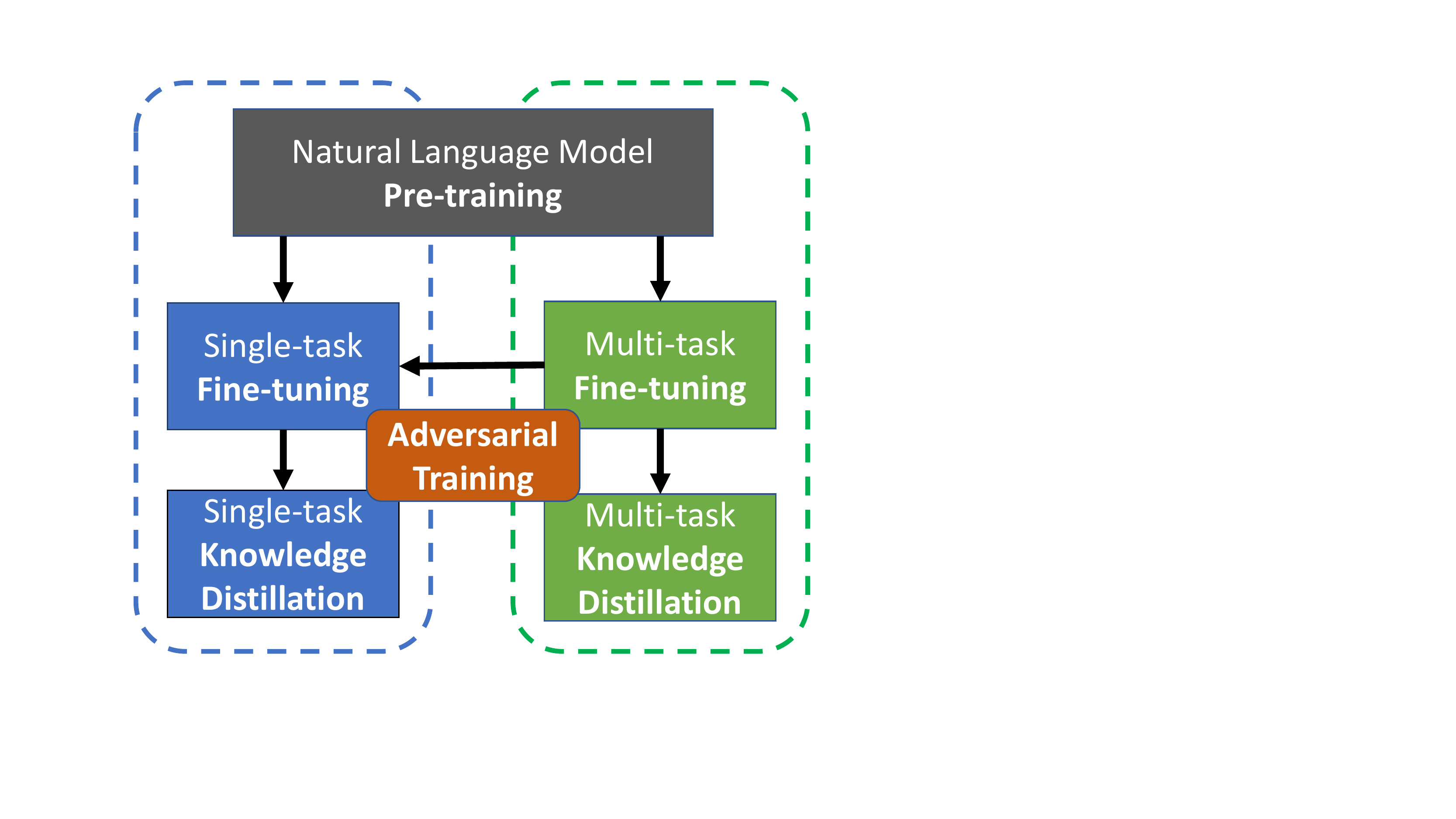}
     }
\vspace{-0.2cm}
	\caption{The workflow of {\DNAME}: train a neural language model on a large amount of unlabeled raw text to obtain general contextual representations; then fine-tune the learned contextual representation on downstream tasks, e.g. GLUE \cite{wang2018glue}; lastly, distill this large model to a lighter one for online deployment. In the later two phrases, we can leverage powerful multi-task learning and adversarial training to further improve performance.}
	\label{fig:workflow}
\vspace{-0.2cm}	
\end{figure}

\begin{figure*}
	\centering
	 \adjustbox{trim={0.0\width} {0.01\height} {0.\width} {0.\height},clip}
     {
 	    \includegraphics[width=0.78\textwidth]{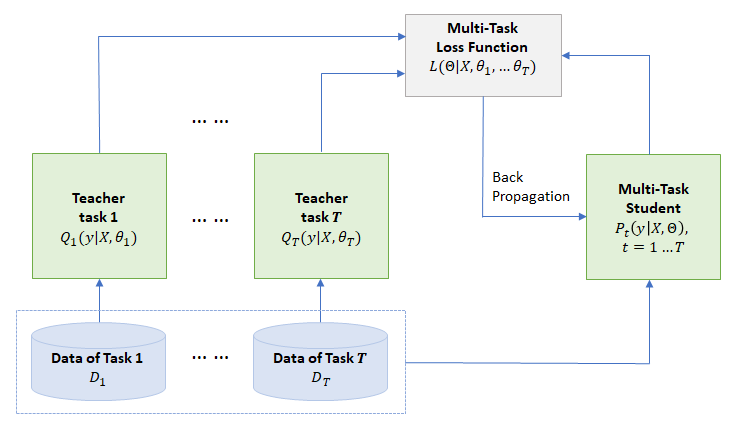}
     }
        \vspace{-0.3cm}
	\caption{Process of knowledge distillation for MTL. A set of tasks where there is task-specific labeled training data are picked. Then, for each task, an ensemble of different neural nets (teacher) is trained. The teacher is used to generate for each task-specific training sample a set of soft targets.  Given the soft targets of the training datasets across multiple tasks, a single MT-DNN (student) shown in Figure~\ref{fig:mt-dnn} is trained using multi-task learning and back propagation, except that if task $t$ has a teacher, the task-specific loss is the average of two objective functions, one for the correct targets and the other for the soft targets assigned by the teacher.
}
	\label{fig:kd}
     \vspace{-0.3cm}
\end{figure*}

\begin{figure*}[ht]
	\centering
     {
 	\includegraphics[width=1.0\textwidth]{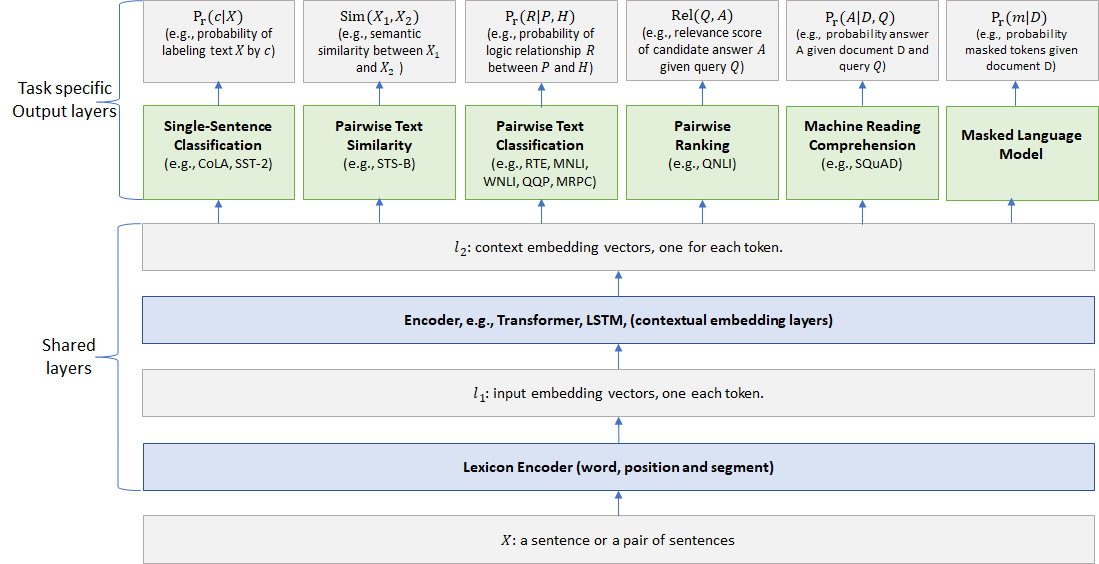}
     }
     \vspace{-0.2cm}
	\caption{Overall System Architecture: The lower layers are shared across all tasks while the top layers are task-specific. The input $X$ (either a sentence or a set of sentences) is first represented as a sequence of embedding vectors, one for each word, in $l_1$. Then the encoder, e.g a Transformer or recurrent neural network (LSTM) model, captures the contextual information for each word and generates the shared contextual embedding vectors in $l_2$. Finally, for each task, additional task-specific layers generate task-specific representations, followed by operations necessary for classification, similarity scoring, or relevance ranking. In case of adversarial training, we perturb embeddings from the lexicon encoder and then add an extra loss term during the training. Note that for the inference phrase, it does not require perturbations.}
	\label{fig:mt-dnn}
     \vspace{-0.4cm}
\end{figure*}

{\DNAME} is designed for modularity, flexibility, and ease of use. These modules are built upon PyTorch \cite{paszke2019pytorch} and Transformers \cite{Wolf2019HuggingFacesTS}, allowing the use of the SOTA pre-trained models, e.g., BERT \cite{devlin2018bert}, RoBERTa \cite{liu2019roberta} and UniLM \cite{dong2019unified}.
The unique attribute of this package is a flexible interface for adversarial multi-task fine-tuning and knowledge distillation, so that researchers and developers can build large SOTA NLU models and then compress them to small ones for online deployment.
The overall workflow and system architecture are shown in Figure~\ref{fig:workflow} and Figure~\ref{fig:mt-dnn} respectively.


\subsection{Workflow}
\vspace{-0.2cm}
As shown in Figure~\ref{fig:workflow}, starting from the neural language model pre-training,
there are three different training configurations by following the directed arrows:
\begin{itemize}
    \vspace{-0.2cm}
    \item Single-task configuration: single-task fine-tuning and single-task knowledge distillation; \vspace{-0.2cm}    
    \item Multi-task configuration: multi-task fine-tuning and multi-task knowledge distillation;
\vspace{-0.3cm} 
   \item Multi-stage configuration: multi-task fine-tuning, single-task fine tuning and single-task knowledge distillation.
\vspace{-0.2cm}
\end{itemize}
Moreover, all configurations can be additionally equipped with the adversarial training.
Each stage of the workflow is described in details as follows.

\noindent \textbf{Neural Language Model Pre-Training}
Due to the great success of deep contextual representations, such as ELMo \cite{peters2018deep}, GPT \cite{gpt22019} and BERT \cite{devlin2018bert}, it is common practice of developing NLU models by first pre-training the underlying neural text representations (text encoders) through massive language modeling
which results in superior text representations transferable across multiple NLP tasks.
Because of this, there has been an increasing effort to develop better pre-trained text encoders by  multiplying either the scale of data \cite{liu2019roberta} or the size of model \cite{raffel2019t5}.
Similar to existing codebases \cite{devlin2018bert}, {\DNAME} supports the LM pre-training from scratch with multiple types of objectives,
such as masked LM \cite{devlin2018bert} and next sentence prediction \cite{devlin2018bert}. 

Moreover, users can leverage the LM pre-training, such as masked LM used by BERT, as an auxiliary task for fine-tuning under the multi-task learning (MTL) framework \cite{sun2019ernie, liu2019mt-dnn}.

\noindent \textbf{Fine-tuning}
Once the text encoder is trained in the pre-training stage, an additional task-specific layer is usually added for fine-tuning based on the downstream task.
Besides the existing typical single-task fine-tuning,
{\DNAME} facilitates a joint fine-tuning with a configurable list of related tasks in a MTL fashion.
By encoding task-relatedness and sharing underlying text representations, MTL is a powerful training paradigm that promotes the model generalization ability and results in improved performance \cite{caruana1997multitask, liu2019mt-dnn, luong2015multi, liu2015mtl,ruder2017overview,collobert2011natural}.
Additionally, a two-step fine-tuning stage is also supported to utilize datasets from related tasks, i.e. a single-task fine-tuning following a multi-task fine-tuning.
It also supports two popular sampling strategies in MTL training:
1) sampling tasks uniformly \cite{caruana1997multitask,liu2015mtl};
2) sampling tasks based on the size of the dataset \cite{liu2019mt-dnn}.
This makes it easy to explore various ways to feed training data to MTL training. 
Finally, to further improve the model robustness,
{\DNAME} also offers a recipe to apply adversarial training \cite{madry2017towards,zhu2019freelb, jiang2019smart} in the fine-tuning stage. 

\noindent \textbf{Knowledge Distillation}
Although contextual text representation models pre-trained with massive text data have led to remarkable progress in NLP,
it is computationally prohibitive and inefficient to deploy such models with millions of parameters for real-world applications (e.g. BERT large model has 344 million parameters).
Therefore, in order to expedite the NLU model learned in either a single-task or multi-task fashion for deployment, {\DNAME} additionally supports the multi-task knowledge distillation  \cite{clark2019bam, liu2019mt-dnn-kd, tang2019distilling,balan2015bayesian,ba2014deep}, an extension of \cite{hinton2015distilling}, to compress cumbersome models into lighter ones. The multi-task knowledge distillation process is illustrated in Figure~\ref{fig:kd}. 
Similar to the fine-tuning stage, adversarial training is available in the knowledge distillation stage.

\subsection{Architecture}
\vspace{-0.2cm}
\paragraph{Lexicon Encoder ($l_1$):} 
The input $X=\{x_1,...,x_m\}$ is a sequence of tokens of length $m$. The first token $x_1$ is always a specific token, e.g. \texttt{[CLS]} for BERT \citet{devlin2018bert} while \texttt{<s>} for RoBERTa \citet{liu2019roberta}. 
If $X$ is a pair of sentences $(X_1, X_2)$, we separate these sentences with special tokens, e.g. \texttt{[SEP]} for BERT and \texttt{[</s>]} for RoBERTa. The lexicon encoder maps $X$ into a sequence of input embedding vectors, one for each token, constructed by summing the corresponding word with positional, and optional segment embeddings. \vspace{-0.2cm}

\paragraph{Encoder ($l_2$):}
We support a multi-layer bidirectional Transformer \cite{vaswani2017attention} or a LSTM \cite{hochreiter1997lstm} encoder to map the input representation vectors ($l_1$) into a sequence of contextual embedding vectors 
$\mathbf{C} \in \mathbb{R}^{d \times m}$. This is the shared representation across different tasks. Note that {\DNAME} also allows developers to customize their own encoders. For example, one can design an encoder with few Transformer layers (e.g. 3 layers) to distill knowledge from the BERT large model (24 layers), so that they can deploy this small mode online to meet the latency restriction as shown in Figure~\ref{fig:kd}.

\paragraph{Task-Specific Output Layers:} 
We can incorporate arbitrary natural language tasks, each with its task-specific output layer. For example, we implement the output layers as a neural decoder for a neural ranker for relevance ranking, a logistic regression for text classification, and so on. A multi-step reasoning decoder, SAN \cite{liu2018san4nli, liu2018san} is also provided.
Customers can choose from existing task-specific output layer or implement new one by themselves.

\section{Application}
In this section, we present a comprehensive set of examples to illustrate how to customize {\DNAME} for new tasks. We use popular benchmarks from general and biomedical domains, including  GLUE \cite{wang2018glue}, SNLI \cite{snli2015}, SciTail \cite{scitail}, SQuAD \cite{squad1}, ANLI \cite{nie2019adversarial}, and biomedical named entity recognition (NER), relation extraction (RE) and question answering (QA) \cite{lee2019biobert}. To make the experiments reproducible, we make all the configuration files publicly available. We also provide a quick guide for customizing a new task in Jupyter notebooks.

\subsection{General Domain Natural Language Understanding Benchmarks}
\begin{table}[!htb]
	\begin{center}
		\begin{tabular}{|@{\hskip1pt}l|l|c|}
			\hline \bf Corpus &Task& Formulation\\ \hline \hline
			\multicolumn{3}{|@{\hskip1pt}c@{\hskip1pt}|}{GLUE} \\ \hline
			CoLA & Acceptability & Classification\\ \hline
			SST & Sentiment& Classification \\ \hline \hline
			MNLI & NLI& Classification\\ \hline
            RTE & NLI & Classification\\ \hline
            WNLI & NLI & Classification\\ \hline
			QQP & Paraphrase& Classification\\ \hline
            MRPC & Paraphrase &Classification\\ \hline
			QNLI & QA/NLI&Classification \\ \hline
			QNLI v1.0 & QA/NLI&Pairwise Ranking \\ \hline
			STS-B & Similarity & Regression\\ \hline
			\multicolumn{3}{|@{\hskip1pt}c@{\hskip1pt}|}{Others} \\ \hline \hline
			SNLI & NLI& Classification\\ \hline
			SciTail & NLI& Classification\\ \hline
			ANLI & NLI& Classification\\ \hline
            SQuAD& MRC& Span Classification\\ \hline
		\end{tabular}
	\end{center}
	\caption{Summary of the four benchmarks: GLUE, SNLI, SciTail and ANLI.
	}
	\label{tab:datasets}
\end{table}

\begin{table}[!htb]
    \centering
    \begin{tabular}{|@{\hskip2pt}l@{\hskip1pt}|@{\hskip1pt}c@{\hskip1pt}|@{\hskip1pt}c@{\hskip1pt}|@{\hskip1pt}c@{\hskip1pt}|@{\hskip1pt}c@{\hskip1pt} |@{\hskip1pt} c@{\hskip1pt}|}
    \hline
   Model                        &MNLI   &RTE    & QNLI  &SST    &MRPC \\ 
                                &Acc    &Acc    &Acc    &Acc    &F1  \\\hline  \hline
     BERT                       &84.5   &63.5   &91.1   &92.9   &89.0 \\ \hline
     BERT + MTL                     &85.3   &79.1   &91.5   &93.6   &89.2 \\ \hline 
     {BERT + AdvTrain}            &85.6   &71.2   &91.6   &93.0   &91.3 \\ \hline \hline
    \end{tabular}
    
    \caption{Comparison among single task, multi-Task and adversarial training on MNLI, RTE, QNLI, SST and MPRC in GLUE. }
    \label{tab:smart_mtl}
\end{table}

\begin{table}[htb!]
	\begin{center}
		\begin{tabular}{| l | c | c|}\hline
	   \bf Model &Dev& Test  \\ \hline 
		BERT\textsubscript{LARGE}  \citep{nie2019adversarial}& 49.3 & 44.2 \\
		\hline
		\hline
		RoBERTa\textsubscript{LARGE} \citep{nie2019adversarial} & 53.7 & 49.7 \\
		\hline 
		RoBERTa-LARGE + AdvTrain  & 57.1 & 57.1 \\ \hline 
 		\end{tabular}
	\end{center}
\vspace{-0.2cm}
\caption{Results in terms of accuracy on the ANLI.}	
	\label{tab:anli}
\end{table}
\noindent $\bullet$ \textbf{GLUE}. The General Language Understanding Evaluation (GLUE) benchmark is a collection of nine natural language understanding (NLU) tasks. As shown in Table~\ref{tab:datasets},
it includes question answering~\cite{squad1}, linguistic acceptability~\cite{cola2018}, sentiment analysis~\cite{sst2013}, text similarity~\cite{sts-b2017}, paraphrase detection~\cite{mrpc2005}, and natural language inference (NLI)~\cite{rte1,rte2,rte3,rte5,winograd2012,mnli2018}. The diversity of the tasks makes GLUE very suitable for evaluating the generalization and robustness of NLU models. 

\noindent $\bullet$ \textbf{SNLI}.
The Stanford Natural Language Inference (SNLI) dataset contains 570k human annotated sentence pairs, in which the premises are drawn from the captions of the Flickr30 corpus and hypotheses are manually annotated \cite{snli2015}. This is the most widely used entailment dataset for NLI.

\noindent $\bullet$ \textbf{SciTail}
This is a textual entailment dataset derived from a science question answering (SciQ) dataset \cite{scitail}. 
In contrast to other entailment datasets mentioned previously, the hypotheses in SciTail are created from science questions while the corresponding answer candidates and premises come from relevant web sentences retrieved from a large corpus. 

\noindent $\bullet$ \textbf{ANLI}.
The Adversarial Natural Language Inference (ANLI, \citet{nie2019adversarial}) is a new large-scale NLI benchmark dataset, collected via an iterative, adversarial human-and-model-in-the-loop procedure. Particular, the data is selected to be difficult to the state-of-the-art models, including BERT and RoBERTa.

\noindent $\bullet$ \textbf{SQuAD}. The Stanford Question Answering Dataset (SQuAD) \cite{squad1} contains about 23K passages and 100K questions. The passages come from approximately 500 Wikipedia articles and the questions and answers are obtained by crowdsourcing. 

Following \cite{devlin2018bert}, table~\ref{tab:smart_mtl} compares different training algorithm: 1) BERT denotes a single task fine-tuning; 2) BERT + MTL indicates that it is trained jointly via MTL; at last 3), BERT + AdvTrain represents that a single task fine-tuning with adversarial training. It is obvious that the both MLT and adversarial training helps to obtain a better result. We further test our model on an adversarial natural language inference (ANLI) dataset \cite{nie2019adversarial}. Table~\ref{tab:anli} summarizes the results on ANLI. As \citet{nie2019adversarial}, all the dataset of ANLI \cite{nie2019adversarial}, MNLI \cite{mnli2018}, SNLI \cite{snli2015} and FEVER \cite{thorne2018fever} are combined as training. {RoBERTa-LARGE+AdvTrain} obtains the best performance compared with all the strong baselines, demonstrating the advantage of adversarial training.

\subsection{Biomedical Natural Language Understating Benchmarks}
\label{subsec:bio}
\vspace{-0.2cm}
There has been rising interest in exploring natural language understanding tasks in high-value domains other than newswire and the Web. In our release, we provide {\DNAME} customization for three representative biomedical natural language understanding tasks:

\noindent $\bullet$ Named entity recognition (NER): In biomedical natural language understanding, NER has received greater attention than other tasks and datasets are available for recognizing various biomedical entities such as disease, gene, drug (chemical).

\noindent $\bullet$ Relation extraction (RE): Relation extraction is more closely related to end applications, but annotation effort is significantly higher compared to NER. Most existing RE tasks focus on binary relations within a short text span such as a sentence of an abstract. Examples include gene-disease or protein-chemical relations. 

\noindent $\bullet$ Question answering (QA): Inspired by interest in QA for the general domain, there has been some effort to create question-answering datasets in biomedicine. Annotation requires domain expertise, so it is significantly harder than in general domain, where it is to produce large-scale datasets by crowdsourcing.

The {\DNAME} customization can work with standard or biomedicine-specific pretraining models such as BioBERT, and can be directly applied to biomedical benchmarks \cite{lee2019biobert}.

\subsection{Extension}
\begin{figure}[ht!]
	 	\centering
	 \adjustbox{trim={0.0\width} {0.01\height} {0.\width} {0.0\height},clip}
     {
 	\includegraphics[width=0.46\textwidth]{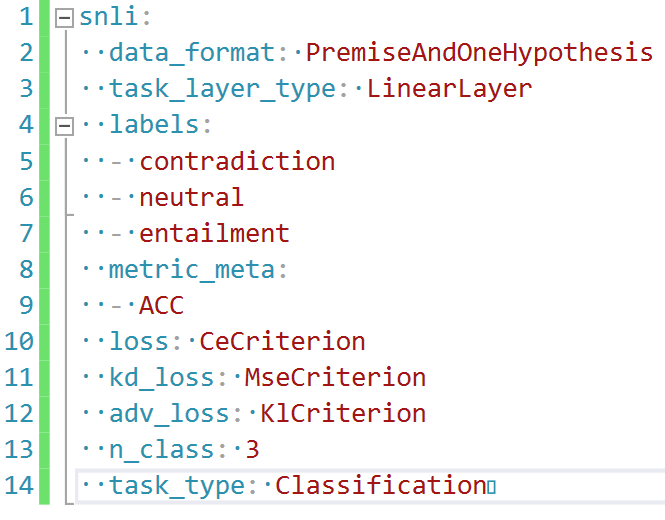}
     }
	\caption{The configuration of SNLI.}
	\label{fig:task}
\end{figure}

We will go though a typical Natural Language Inference task, e.g. SNLI, which is one of the most popular benchmark, showing how to apply our toolkit to a new task.  {\DNAME} is driven by configuration and command line arguments. Firstly, the SNLI configuration is shown in Figure~\ref{fig:task}. The configuration defines tasks, model architecture as well as loss functions. We briefly introduce these attributes as follows: 
\begin{enumerate}
    \vspace{-0.2cm}
    \item \textit{data\_format} is a required attribute and it denotes that each sample includes two sentences (premise and hypothesis). Please refer the tutorial and API for supported formats.
    \vspace{-0.2cm}
    \item \textit{task\_layer\_type} specifies architecture of the task specific layer. The default is a "linear layer". 
    \vspace{-0.2cm}
    \item \textit{labels} Users can list unique values of labels. The configuration helps to convert back and forth between text labels and numbers during training and evaluation. Without it, {\DNAME} assumes the label of prediction are numbers.
    \vspace{-0.2cm}
    \item \textit{metric\_meta} is the evaluation metric used for validation. 
    \item \textit{loss} is the loss function for SNLI. It also supports other functions, e.g. MSE for regression.
    \vspace{-0.2cm}
    
    \item \textit{kd\_loss} is the loss function in the knowledge distillation setting.
    \vspace{-0.2cm}
    \item \textit{adv\_loss} is the loss function in the adversarial setting.
    \vspace{-0.2cm}
    \item \textit{n\_class} denotes the number of categories for SNLI.
    \vspace{-0.2cm}
    \item \textit{task\_type} specifies whether it is a classification task or a regression task.
    \vspace{-0.2cm}
\end{enumerate}

Once the configuration is provided, one can train the customized model for the task, using any supported pre-trained models as starting point.

{\DNAME} is also highly extensible, as shown in Figure~\ref{fig:task}, \textit{loss} and \textit{task\_layer\_type} point to existing classes in code. Users can write customized classes and plug into {\DNAME}. The customized classes could then be used via configuration.


\section{Conclusion}
\label{sec:conclusion}
Microsoft {\DNAME} is an open-source natural language understanding toolkit which facilitates researchers and developers to build customized deep learning models.
Its key features are: 1) support for robust and transferable learning using adversarial multi-task learning paradigm; 2) 
enable knowledge distillation under the multi-task learning setting which can be leveraged to derive lighter models for efficient online deployment.
We will extend {\DNAME} to support Natural Language Generation tasks, e.g. Question Generation, and incorporate more pre-trained encoders, e.g. T5 \cite{raffel2019t5} in future.  
\section*{Acknowledgments}
\vspace{-2mm}
We thank Liyuan Liu, Sha Li, Mehrad Moradshahi and other contributors to the package, and the anonymous reviewers for valuable discussions and comments. 
\bibliography{ref}

\begin{thebibliography}{51}
\expandafter\ifx\csname natexlab\endcsname\relax\def\natexlab#1{#1}\fi

\bibitem[{Ba and Caruana(2014)}]{ba2014deep}
Jimmy Ba and Rich Caruana. 2014.
\newblock Do deep nets really need to be deep?
\newblock In \emph{Advances in neural information processing systems}, pages
  2654--2662.

\bibitem[{Balan et~al.(2015)Balan, Rathod, Murphy, and
  Welling}]{balan2015bayesian}
Anoop~Korattikara Balan, Vivek Rathod, Kevin~P Murphy, and Max Welling. 2015.
\newblock Bayesian dark knowledge.
\newblock In \emph{Advances in Neural Information Processing Systems}, pages
  3438--3446.

\bibitem[{Bar-Haim et~al.(2006)Bar-Haim, Dagan, Dolan, Ferro, and
  Giampiccolo}]{rte2}
Roy Bar-Haim, Ido Dagan, Bill Dolan, Lisa Ferro, and Danilo Giampiccolo. 2006.
\newblock The second {PASCAL} recognising textual entailment challenge.
\newblock In \emph{Proceedings of the Second {PASCAL} Challenges Workshop on
  Recognising Textual Entailment}.

\bibitem[{Bentivogli et~al.(2009)Bentivogli, Dagan, Dang, Giampiccolo, and
  Magnini}]{rte5}
Luisa Bentivogli, Ido Dagan, Hoa~Trang Dang, Danilo Giampiccolo, and Bernardo
  Magnini. 2009.
\newblock The fifth pascal recognizing textual entailment challenge.
\newblock In \emph{In Proc Text Analysis Conference (TAC’09}.

\bibitem[{Bowman et~al.(2015)Bowman, Angeli, Potts, and Manning}]{snli2015}
Samuel~R. Bowman, Gabor Angeli, Christopher Potts, and Christopher~D. Manning.
  2015.
\newblock A large annotated corpus for learning natural language inference.
\newblock In \emph{Proceedings of the 2015 Conference on Empirical Methods in
  Natural Language Processing (EMNLP)}. Association for Computational
  Linguistics.

\bibitem[{Caruana(1997)}]{caruana1997multitask}
Rich Caruana. 1997.
\newblock Multitask learning.
\newblock \emph{Machine learning}, 28(1):41--75.

\bibitem[{Cer et~al.(2017)Cer, Diab, Agirre, Lopez-Gazpio, and
  Specia}]{sts-b2017}
Daniel Cer, Mona Diab, Eneko Agirre, Inigo Lopez-Gazpio, and Lucia Specia.
  2017.
\newblock Semeval-2017 task 1: Semantic textual similarity-multilingual and
  cross-lingual focused evaluation.
\newblock \emph{arXiv preprint arXiv:1708.00055}.

\bibitem[{Clark et~al.(2019)Clark, Luong, Khandelwal, Manning, and
  Le}]{clark2019bam}
Kevin Clark, Minh-Thang Luong, Urvashi Khandelwal, Christopher~D Manning, and
  Quoc~V Le. 2019.
\newblock Bam! born-again multi-task networks for natural language
  understanding.
\newblock \emph{arXiv preprint arXiv:1907.04829}.

\bibitem[{Collobert et~al.(2011)Collobert, Weston, Bottou, Karlen, Kavukcuoglu,
  and Kuksa}]{collobert2011natural}
Ronan Collobert, Jason Weston, L{\'e}on Bottou, Michael Karlen, Koray
  Kavukcuoglu, and Pavel Kuksa. 2011.
\newblock Natural language processing (almost) from scratch.
\newblock \emph{Journal of machine learning research}, 12(Aug):2493--2537.

\bibitem[{Dagan et~al.(2006)Dagan, Glickman, and Magnini}]{rte1}
Ido Dagan, Oren Glickman, and Bernardo Magnini. 2006.
\newblock \href {https://doi.org/10.1007/11736790_9} {The pascal recognising
  textual entailment challenge}.
\newblock In \emph{Proceedings of the First International Conference on Machine
  Learning Challenges: Evaluating Predictive Uncertainty Visual Object
  Classification, and Recognizing Textual Entailment}, MLCW'05, pages 177--190,
  Berlin, Heidelberg. Springer-Verlag.

\bibitem[{Devlin et~al.(2019)Devlin, Chang, Lee, and
  Toutanova}]{devlin2018bert}
Jacob Devlin, Ming-Wei Chang, Kenton Lee, and Kristina Toutanova. 2019.
\newblock Bert: Pre-training of deep bidirectional transformers for language
  understanding.
\newblock In \emph{Proceedings of the 2019 Conference of the North American
  Chapter of the Association for Computational Linguistics: Human Language
  Technologies, Volume 1 (Long and Short Papers)}, pages 4171--4186.

\bibitem[{Dolan and Brockett(2005)}]{mrpc2005}
William~B Dolan and Chris Brockett. 2005.
\newblock Automatically constructing a corpus of sentential paraphrases.
\newblock In \emph{Proceedings of the Third International Workshop on
  Paraphrasing (IWP2005)}.

\bibitem[{Dong et~al.(2019)Dong, Yang, Wang, Wei, Liu, Wang, Gao, Zhou, and
  Hon}]{dong2019unified}
Li~Dong, Nan Yang, Wenhui Wang, Furu Wei, Xiaodong Liu, Yu~Wang, Jianfeng Gao,
  Ming Zhou, and Hsiao-Wuen Hon. 2019.
\newblock Unified language model pre-training for natural language
  understanding and generation.
\newblock In \emph{Advances in Neural Information Processing Systems}, pages
  13042--13054.

\bibitem[{Gardner et~al.(2018)Gardner, Grus, Neumann, Tafjord, Dasigi, Liu,
  Peters, Schmitz, and Zettlemoyer}]{gardner2018allennlp}
Matt Gardner, Joel Grus, Mark Neumann, Oyvind Tafjord, Pradeep Dasigi, Nelson
  Liu, Matthew Peters, Michael Schmitz, and Luke Zettlemoyer. 2018.
\newblock Allennlp: A deep semantic natural language processing platform.
\newblock \emph{arXiv preprint arXiv:1803.07640}.

\bibitem[{Giampiccolo et~al.(2007)Giampiccolo, Magnini, Dagan, and
  Dolan}]{rte3}
Danilo Giampiccolo, Bernardo Magnini, Ido Dagan, and Bill Dolan. 2007.
\newblock \href {https://www.aclweb.org/anthology/W07-1401} {The third {PASCAL}
  recognizing textual entailment challenge}.
\newblock In \emph{Proceedings of the {ACL}-{PASCAL} Workshop on Textual
  Entailment and Paraphrasing}, pages 1--9, Prague. Association for
  Computational Linguistics.

\bibitem[{Hinton et~al.(2015)Hinton, Vinyals, and Dean}]{hinton2015distilling}
Geoffrey Hinton, Oriol Vinyals, and Jeff Dean. 2015.
\newblock Distilling the knowledge in a neural network.
\newblock \emph{arXiv preprint arXiv:1503.02531}.

\bibitem[{Hochreiter and Schmidhuber(1997)}]{hochreiter1997lstm}
Sepp Hochreiter and J{\"u}rgen Schmidhuber. 1997.
\newblock Long short-term memory.
\newblock \emph{Neural computation}, 9(8):1735--1780.

\bibitem[{Jiang et~al.(2019)Jiang, He, Chen, Liu, Gao, and
  Zhao}]{jiang2019smart}
Haoming Jiang, Pengcheng He, Weizhu Chen, Xiaodong Liu, Jianfeng Gao, and Tuo
  Zhao. 2019.
\newblock Smart: Robust and efficient fine-tuning for pre-trained natural
  language models through principled regularized optimization.
\newblock \emph{arXiv preprint arXiv:1911.03437}.

\bibitem[{Khot et~al.(2018)Khot, Sabharwal, and Clark}]{scitail}
Tushar Khot, Ashish Sabharwal, and Peter Clark. 2018.
\newblock {SciTail}: A textual entailment dataset from science question
  answering.
\newblock In \emph{AAAI}.

\bibitem[{Lee et~al.(2019)Lee, Yoon, Kim, Kim, Kim, So, and
  Kang}]{lee2019biobert}
Jinhyuk Lee, Wonjin Yoon, Sungdong Kim, Donghyeon Kim, Sunkyu Kim, Chan~Ho So,
  and Jaewoo Kang. 2019.
\newblock Biobert: pre-trained biomedical language representation model for
  biomedical text mining.
\newblock \emph{arXiv preprint arXiv:1901.08746}.

\bibitem[{Levesque et~al.(2012)Levesque, Davis, and Morgenstern}]{winograd2012}
Hector Levesque, Ernest Davis, and Leora Morgenstern. 2012.
\newblock The winograd schema challenge.
\newblock In \emph{Thirteenth International Conference on the Principles of
  Knowledge Representation and Reasoning}.

\bibitem[{Liu et~al.(2020)Liu, Cheng, He, Chen, Wang, Poon, and
  Gao}]{liu2020alum}
Xiaodong Liu, Hao Cheng, Pengcheng He, Weizhu Chen, Yu~Wang, Hoifung Poon, and
  Jianfeng Gao. 2020.
\newblock Adversarial training for large neural language models.
\newblock \emph{arXiv preprint arXiv:2004.08994}.

\bibitem[{Liu et~al.(2018{\natexlab{a}})Liu, Duh, and Gao}]{liu2018san4nli}
Xiaodong Liu, Kevin Duh, and Jianfeng Gao. 2018{\natexlab{a}}.
\newblock Stochastic answer networks for natural language inference.
\newblock \emph{arXiv preprint arXiv:1804.07888}.

\bibitem[{Liu et~al.(2015)Liu, Gao, He, Deng, Duh, and Wang}]{liu2015mtl}
Xiaodong Liu, Jianfeng Gao, Xiaodong He, Li~Deng, Kevin Duh, and Ye-Yi Wang.
  2015.
\newblock Representation learning using multi-task deep neural networks for
  semantic classification and information retrieval.
\newblock In \emph{Proceedings of the 2015 Conference of the North American
  Chapter of the Association for Computational Linguistics: Human Language
  Technologies}, pages 912--921.

\bibitem[{Liu et~al.(2019{\natexlab{a}})Liu, He, Chen, and
  Gao}]{liu2019mt-dnn-kd}
Xiaodong Liu, Pengcheng He, Weizhu Chen, and Jianfeng Gao. 2019{\natexlab{a}}.
\newblock Improving multi-task deep neural networks via knowledge distillation
  for natural language understanding.
\newblock \emph{arXiv preprint arXiv:1904.09482}.

\bibitem[{Liu et~al.(2019{\natexlab{b}})Liu, He, Chen, and Gao}]{liu2019mt-dnn}
Xiaodong Liu, Pengcheng He, Weizhu Chen, and Jianfeng Gao. 2019{\natexlab{b}}.
\newblock \href {https://www.aclweb.org/anthology/P19-1441} {Multi-task deep
  neural networks for natural language understanding}.
\newblock In \emph{Proceedings of the 57th Annual Meeting of the Association
  for Computational Linguistics}, pages 4487--4496, Florence, Italy.
  Association for Computational Linguistics.

\bibitem[{Liu et~al.(2018{\natexlab{b}})Liu, Shen, Duh, and Gao}]{liu2018san}
Xiaodong Liu, Yelong Shen, Kevin Duh, and Jianfeng Gao. 2018{\natexlab{b}}.
\newblock Stochastic answer networks for machine reading comprehension.
\newblock In \emph{Proceedings of the 56th Annual Meeting of the Association
  for Computational Linguistics (Volume 1: Long Papers)}. Association for
  Computational Linguistics.

\bibitem[{Liu et~al.(2019{\natexlab{c}})Liu, Ott, Goyal, Du, Joshi, Chen, Levy,
  Lewis, Zettlemoyer, and Stoyanov}]{liu2019roberta}
Yinhan Liu, Myle Ott, Naman Goyal, Jingfei Du, Mandar Joshi, Danqi Chen, Omer
  Levy, Mike Lewis, Luke Zettlemoyer, and Veselin Stoyanov. 2019{\natexlab{c}}.
\newblock Roberta: A robustly optimized bert pretraining approach.
\newblock \emph{arXiv preprint arXiv:1907.11692}.

\bibitem[{Luong et~al.(2015)Luong, Le, Sutskever, Vinyals, and
  Kaiser}]{luong2015multi}
Minh-Thang Luong, Quoc~V Le, Ilya Sutskever, Oriol Vinyals, and Lukasz Kaiser.
  2015.
\newblock Multi-task sequence to sequence learning.
\newblock \emph{arXiv preprint arXiv:1511.06114}.

\bibitem[{Madry et~al.(2017)Madry, Makelov, Schmidt, Tsipras, and
  Vladu}]{madry2017towards}
Aleksander Madry, Aleksandar Makelov, Ludwig Schmidt, Dimitris Tsipras, and
  Adrian Vladu. 2017.
\newblock Towards deep learning models resistant to adversarial attacks.
\newblock \emph{arXiv preprint arXiv:1706.06083}.

\bibitem[{Manning et~al.(2014)Manning, Surdeanu, Bauer, Finkel, Bethard, and
  McClosky}]{manning2014stanford}
Christopher~D Manning, Mihai Surdeanu, John Bauer, Jenny~Rose Finkel, Steven
  Bethard, and David McClosky. 2014.
\newblock The stanford corenlp natural language processing toolkit.
\newblock In \emph{Proceedings of 52nd annual meeting of the association for
  computational linguistics: system demonstrations}, pages 55--60.

\bibitem[{Miyato et~al.(2018)Miyato, Maeda, Koyama, and
  Ishii}]{miyato2018virtual}
Takeru Miyato, Shin-ichi Maeda, Masanori Koyama, and Shin Ishii. 2018.
\newblock Virtual adversarial training: a regularization method for supervised
  and semi-supervised learning.
\newblock \emph{IEEE transactions on pattern analysis and machine
  intelligence}, 41(8):1979--1993.

\bibitem[{Nie et~al.(2019)Nie, Williams, Dinan, Bansal, Weston, and
  Kiela}]{nie2019adversarial}
Yixin Nie, Adina Williams, Emily Dinan, Mohit Bansal, Jason Weston, and Douwe
  Kiela. 2019.
\newblock Adversarial nli: A new benchmark for natural language understanding.
\newblock \emph{arXiv preprint arXiv:1910.14599}.

\bibitem[{Ott et~al.(2019)Ott, Edunov, Baevski, Fan, Gross, Ng, Grangier, and
  Auli}]{ott2019fairseq}
Myle Ott, Sergey Edunov, Alexei Baevski, Angela Fan, Sam Gross, Nathan Ng,
  David Grangier, and Michael Auli. 2019.
\newblock fairseq: A fast, extensible toolkit for sequence modeling.
\newblock \emph{arXiv preprint arXiv:1904.01038}.

\bibitem[{Paszke et~al.(2019)Paszke, Gross, Massa, Lerer, Bradbury, Chanan,
  Killeen, Lin, Gimelshein, Antiga et~al.}]{paszke2019pytorch}
Adam Paszke, Sam Gross, Francisco Massa, Adam Lerer, James Bradbury, Gregory
  Chanan, Trevor Killeen, Zeming Lin, Natalia Gimelshein, Luca Antiga, et~al.
  2019.
\newblock Pytorch: An imperative style, high-performance deep learning library.
\newblock In \emph{Advances in Neural Information Processing Systems}, pages
  8024--8035.

\bibitem[{Peters et~al.(2018)Peters, Neumann, Iyyer, Gardner, Clark, Lee, and
  Zettlemoyer}]{peters2018deep}
Matthew~E Peters, Mark Neumann, Mohit Iyyer, Matt Gardner, Christopher Clark,
  Kenton Lee, and Luke Zettlemoyer. 2018.
\newblock Deep contextualized word representations.
\newblock \emph{arXiv preprint arXiv:1802.05365}.

\bibitem[{Radford et~al.(2018)Radford, Wu, Child, Luan, Amodei, and
  Sutskever}]{gpt22019}
Alec Radford, Jeffrey Wu, Rewon Child, David Luan, Dario Amodei, and Ilya
  Sutskever. 2018.
\newblock Language models are unsupervised multitask learners.

\bibitem[{Raffel et~al.(2019)Raffel, Shazeer, Roberts, Lee, Narang, Matena,
  Zhou, Li, and Liu}]{raffel2019t5}
Colin Raffel, Noam Shazeer, Adam Roberts, Katherine Lee, Sharan Narang, Michael
  Matena, Yanqi Zhou, Wei Li, and Peter~J Liu. 2019.
\newblock Exploring the limits of transfer learning with a unified text-to-text
  transformer.
\newblock \emph{arXiv preprint arXiv:1910.10683}.

\bibitem[{Rajpurkar et~al.(2016)Rajpurkar, Zhang, Lopyrev, and Liang}]{squad1}
Pranav Rajpurkar, Jian Zhang, Konstantin Lopyrev, and Percy Liang. 2016.
\newblock \href {https://doi.org/10.18653/v1/D16-1264} {{SQ}u{AD}: 100,000+
  questions for machine comprehension of text}.
\newblock In \emph{Proceedings of the 2016 Conference on Empirical Methods in
  Natural Language Processing}, pages 2383--2392, Austin, Texas. Association
  for Computational Linguistics.

\bibitem[{Ruder(2017)}]{ruder2017overview}
Sebastian Ruder. 2017.
\newblock An overview of multi-task learning in deep neural networks.
\newblock \emph{arXiv preprint arXiv:1706.05098}.

\bibitem[{Sanh et~al.(2019)Sanh, Debut, Chaumond, and
  Wolf}]{sanh2019distilbert}
Victor Sanh, Lysandre Debut, Julien Chaumond, and Thomas Wolf. 2019.
\newblock Distilbert, a distilled version of bert: smaller, faster, cheaper and
  lighter.
\newblock \emph{arXiv preprint arXiv:1910.01108}.

\bibitem[{Socher et~al.(2013)Socher, Perelygin, Wu, Chuang, Manning, Ng, and
  Potts}]{sst2013}
Richard Socher, Alex Perelygin, Jean Wu, Jason Chuang, Christopher~D Manning,
  Andrew Ng, and Christopher Potts. 2013.
\newblock Recursive deep models for semantic compositionality over a sentiment
  treebank.
\newblock In \emph{Proceedings of the 2013 conference on empirical methods in
  natural language processing}, pages 1631--1642.

\bibitem[{Sun et~al.(2019)Sun, Wang, Li, Feng, Tian, Wu, and
  Wang}]{sun2019ernie}
Yu~Sun, Shuohuan Wang, Yukun Li, Shikun Feng, Hao Tian, Hua Wu, and Haifeng
  Wang. 2019.
\newblock Ernie 2.0: A continual pre-training framework for language
  understanding.
\newblock \emph{arXiv preprint arXiv:1907.12412}.

\bibitem[{Tang et~al.(2019)Tang, Lu, Liu, Mou, Vechtomova, and
  Lin}]{tang2019distilling}
Raphael Tang, Yao Lu, Linqing Liu, Lili Mou, Olga Vechtomova, and Jimmy Lin.
  2019.
\newblock Distilling task-specific knowledge from bert into simple neural
  networks.
\newblock \emph{arXiv preprint arXiv:1903.12136}.

\bibitem[{Thorne et~al.(2018)Thorne, Vlachos, Christodoulopoulos, and
  Mittal}]{thorne2018fever}
James Thorne, Andreas Vlachos, Christos Christodoulopoulos, and Arpit Mittal.
  2018.
\newblock Fever: a large-scale dataset for fact extraction and verification.
\newblock \emph{arXiv preprint arXiv:1803.05355}.

\bibitem[{Vaswani et~al.(2017)Vaswani, Shazeer, Parmar, Uszkoreit, Jones,
  Gomez, Kaiser, and Polosukhin}]{vaswani2017attention}
Ashish Vaswani, Noam Shazeer, Niki Parmar, Jakob Uszkoreit, Llion Jones,
  Aidan~N Gomez, {\L}ukasz Kaiser, and Illia Polosukhin. 2017.
\newblock Attention is all you need.
\newblock In \emph{Advances in neural information processing systems}, pages
  5998--6008.

\bibitem[{Wang et~al.(2018)Wang, Singh, Michael, Hill, Levy, and
  Bowman}]{wang2018glue}
Alex Wang, Amanpreet Singh, Julian Michael, Felix Hill, Omer Levy, and Samuel~R
  Bowman. 2018.
\newblock Glue: A multi-task benchmark and analysis platform for natural
  language understanding.
\newblock \emph{arXiv preprint arXiv:1804.07461}.

\bibitem[{Warstadt et~al.(2018)Warstadt, Singh, and Bowman}]{cola2018}
Alex Warstadt, Amanpreet Singh, and Samuel~R Bowman. 2018.
\newblock Neural network acceptability judgments.
\newblock \emph{arXiv preprint arXiv:1805.12471}.

\bibitem[{Williams et~al.(2018)Williams, Nangia, and Bowman}]{mnli2018}
Adina Williams, Nikita Nangia, and Samuel Bowman. 2018.
\newblock \href {http://aclweb.org/anthology/N18-1101} {A broad-coverage
  challenge corpus for sentence understanding through inference}.
\newblock In \emph{Proceedings of the 2018 Conference of the North American
  Chapter of the Association for Computational Linguistics: Human Language
  Technologies, Volume 1 (Long Papers)}, pages 1112--1122. Association for
  Computational Linguistics.

\bibitem[{Wolf et~al.(2019)Wolf, Debut, Sanh, Chaumond, Delangue, Moi, Cistac,
  Rault, Louf, Funtowicz, and Brew}]{Wolf2019HuggingFacesTS}
Thomas Wolf, Lysandre Debut, Victor Sanh, Julien Chaumond, Clement Delangue,
  Anthony Moi, Pierric Cistac, Tim Rault, R'emi Louf, Morgan Funtowicz, and
  Jamie Brew. 2019.
\newblock Huggingface's transformers: State-of-the-art natural language
  processing.
\newblock \emph{ArXiv}, abs/1910.03771.

\bibitem[{Zhu et~al.(2019)Zhu, Cheng, Gan, Sun, Goldstein, and
  Liu}]{zhu2019freelb}
Chen Zhu, Yu~Cheng, Zhe Gan, Siqi Sun, Thomas Goldstein, and Jingjing Liu.
  2019.
\newblock Freelb: Enhanced adversarial training for language understanding.
\newblock \emph{arXiv preprint arXiv:1909.11764}.

\end{thebibliography}
\bibliographystyle{acl_natbib}
\end{document}